\documentclass[letterpaper, 10 pt, conference]{ieeeconf}
\IEEEoverridecommandlockouts

\newcommand\defeq{\mathrel{\stackrel{\makebox[0pt]{\mbox{\normalfont\scriptsize def}}}{:=\,}}}
\usepackage{amsmath}
\usepackage{amssymb}
\usepackage{hyperref}
\usepackage{graphicx}
\usepackage{multirow}
\usepackage[table,xcdraw]{xcolor}
\usepackage{arydshln}
\usepackage{caption}
\usepackage{balance}

\usepackage{tikz}
\newcommand\copyrighttext{%
	\footnotesize © 2024 IEEE.  Personal use of this material is permitted.  Permission from IEEE must be obtained for all other uses, in any current or future media, including reprinting/republishing this material for advertising or promotional purposes, creating new collective works, for resale or redistribution to servers or lists, or reuse of any copyrighted component of this work in other works.}
\newcommand\copyrightnotice{%
	\begin{tikzpicture}[remember picture,overlay]
		\node[anchor=south,yshift=10pt] at (current page.south) {\fbox{\parbox{\dimexpr\textwidth-\fboxsep-\fboxrule\relax}{\copyrighttext}}};
	\end{tikzpicture}%
}

\begin{document}
	
	\title{\LARGE \bf 
		USC: Uncompromising Spatial Constraints for Safety-Oriented\\3D Object Detectors in Autonomous Driving
	}
	
	\author{Brian Hsuan-Cheng Liao$^{\dagger\ddagger}$, Chih-Hong Cheng$^{\mathsection\sharp}$, Hasan Esen$^{\dagger}$, Alois Knoll$^{\ddagger}$ %
		\thanks{$^{\dagger}$DENSO AUTOMOTIVE Deutschland GmbH, Germany} %
		\thanks{$^{\ddagger}$Technical University of Munich, Germany} %
		\thanks{$^{\mathsection}$Chalmers University of Technology, Sweden} %
		\thanks{$^{\sharp}$University of Gothenburg, Sweden} %
		\thanks{Correspondence to \tt\small {h.liao}@eu.denso.com}
	}
		
	\maketitle
	\copyrightnotice

	\begin{abstract}
		
		In this work, we consider the safety-oriented performance of 3D object detectors in autonomous driving contexts. Specifically, despite impressive results shown by the mass literature, developers often find it hard to ensure the safe deployment of these learning-based perception models. Attributing the challenge to the lack of safety-oriented metrics, we hereby present uncompromising spatial constraints (USC), which characterize a simple yet important localization requirement demanding the predictions to fully cover the objects when seen from the autonomous vehicle. The constraints, as we formulate using the perspective and bird's-eye views, can be naturally reflected by quantitative measures, such that having an object detector with a higher score implies a lower risk of collision. Finally, beyond model evaluation, we incorporate the quantitative measures into common loss functions to enable safety-oriented fine-tuning for existing models. With experiments using the nuScenes dataset and a closed-loop simulation, our work demonstrates such considerations of safety notions at the perception level not only improve model performances beyond accuracy but also allow for a more direct linkage to actual system safety.
		
	\end{abstract}

	\section{Introduction}
	\label{sec:intro}
	
	3D object detection is essential for scene understanding in autonomous driving (AD). It allows an autonomous vehicle (AV) to locate objects in a scene and classify them into different categories so that the AV can navigate through them effectively. 
	
	In the past years, computer vision and deep learning literature have contributed greatly to the growing performances of 3D object detectors. More recently, aiming at broader and safer deployment of AVs, several works have adapted the general literature and designed performance indicators that consider driving dynamics. For example, instead of the generic Intersection-over-Union (IoU) measure~\cite{geiger2012are}, the nuScenes benchmark uses translation errors (TE) to match the predictions and ground truths and calculates several true-positive metrics for more fine-grained evaluation~\cite{caesar2020nuscenes}. 
	While various driving-oriented metrics are proposed, a remaining challenge is the definition of a straightforward criterion, i.e., what score a model should attain to be satisfactory and deployable. Lately, the mitigation of the challenge has become more urgent, as regulations such as the EU AI Act~\cite{aiact} require developers to demonstrate ``an appropriate level of performance" before marketing their solutions.

	\begin{figure}
		\centering
		\includegraphics[width=0.95\linewidth]{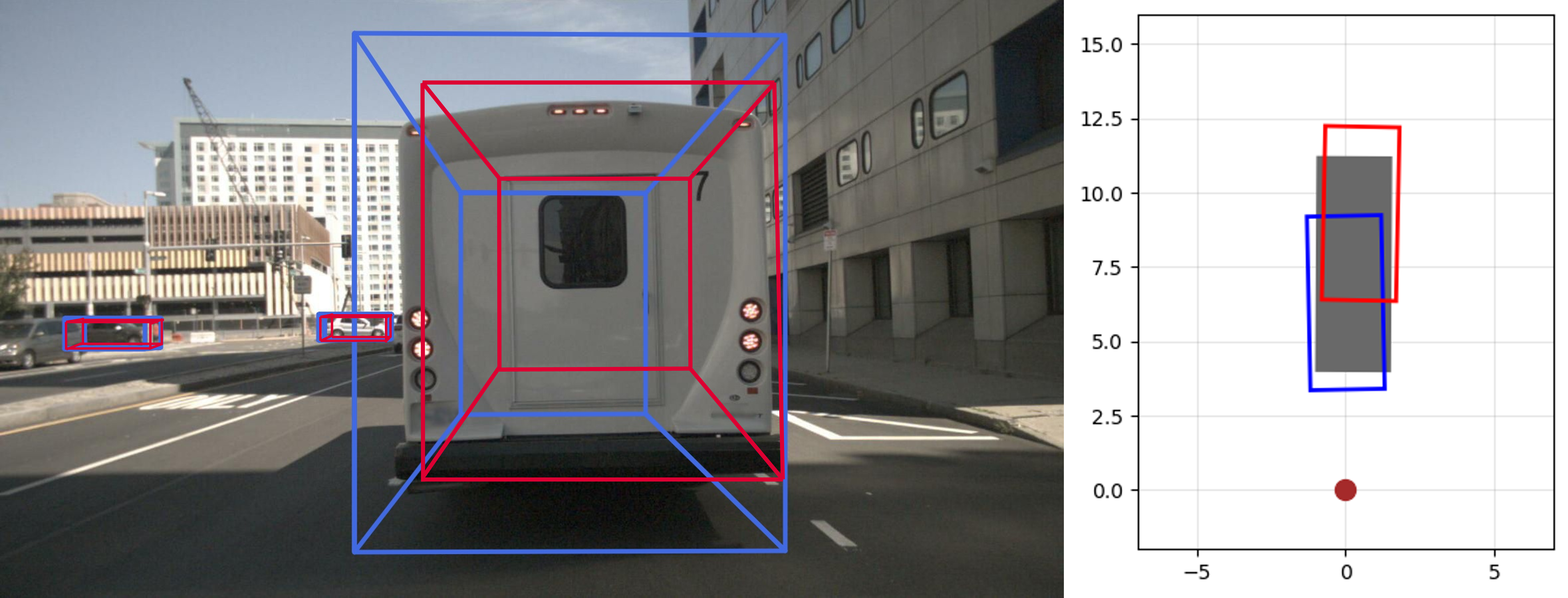}
		\caption{An example scene that motivates our safety-oriented spatial constraints. For both the blue and the red predictions, $\mathsf{IoU}\approx0.7$ and $\mathsf{TE}\approx1.6 \, (m)$ indistinguishably. We thereby focus on the objects' nearest sides from the AV and propose USC. Effectively, the blue prediction has $\mathsf{USC}=1$, while the red one gets $\mathsf{USC}\approx0.8$.}
		\label{fig:example}
	\end{figure}
	
	Hence, in this work, we consider tolerable errors for 3D object detectors from a safety point of view and propose uncompromising spatial constraints (USC) as potential bottom lines. Concretely, \emph{pivoting on object response and collision avoidance}, we underpin a requirement as follows: An object detector's prediction should fully cover its corresponding ground truth when seen from the AV's location. The reason is that such full coverage should already result in a low risk of collision at the object's edges close to the AV. Fig.~\ref{fig:example} shows an example scene and contrasts USC with existing practices, which generally demand perfect alignment between the prediction and the ground truth.
	
	More technically, when characterizing the requirement, we utilize perspective-view (PV) and bird's-eye-view (BEV) projections to decompose the 3D predictions and ground truths into 2D representations. Such decomposition facilitates the definition of USC and, more importantly, leads to quantitative measures that allow us to assess different models at a finer scale. Notably, while the PV typically captures rich information for AD as shown by the prior art~\cite{degrancey2022object,schuster2022unaligned}, we point out the necessity of BEV constraints for the required full coverage. Moreover, beyond assessment, we exploit the quantitative measures and formulate a safety-oriented loss function to improve existing models.
	
	To demonstrate the efficacy of our proposals, we conduct three sets of experiments with state-of-the-art 3D object detectors. First, we evaluate the models using the nuScenes dataset~\cite{caesar2020nuscenes} and check for safety concerns via USC-based metrics\footnote{The utilization of the nuScenes datasets in this paper is for knowledge dissemination and scientific publication and is not for commercial use.}. We then improve one of the models through the safety-oriented loss function. Finally, by running the object detectors with an automatic emergency braking (AEB) function in closed-loop simulations, we show that USC enables a better correlation between component-level metrics and the AV's collision rates. Altogether, our work serves as an initiative to address safety principles and remedies for imperfect 3D object detectors in autonomous driving systems.

	\section{Related Work}
	\label{sec:related_work}
	
	Object detection has attracted great research attention over the past two decades. Readers may refer to a recent survey that delineates the development~\cite{zou2023object}. In the following section, we focus on studies that are concerned with AD and safety. 
	
	As one of the early results, Cheng et al. proposed a conceptual safety-aware hardening framework for object detectors~\cite{cheng2020safety}. Specifically, the authors defined a distance-based criticality criterion and adapted the NN architecture of a baseline model to enhance its performance in critical zones. Based on a similar idea, Lyssenko et al. refined the definition of criticality by considering object dynamics and improved pedestrian detection accordingly~\cite{lyssenko2024safety}. While these results contributed to better recognition of safety-critical objects in general, our study complements them by precising localization requirements at a finer scale. In fact, the introduced nuScenes benchmark lies in this direction~\cite{caesar2020nuscenes}. It adopts the NuScenes Detection Score (NDS), computing not only the conventional mean average precision (mAP) metric~\cite{zou2023object} but also true-positive error measures on the localization function. Concurrently, Philion et al. proposed planner-centric metrics, which compute the KL-divergence between the motion planning outcomes (i.e., trajectory waypoints) based on predictions and ground truths respectively~\cite{philion2020learning}. Although the latter approach appears plausible, a recent validation work interestingly found NDS better correlating with actual AV collision rates, indicating the advantage of true-positive error measures at the object detection model level~\cite{schreier2023offline}. Hence, our investigation follows this direction, but as we shall see, while NDS demands perfectly aligned predictions, our work focuses on safety-oriented principles.
	
	In the most recent literature, several similar efforts have been made. For instance, Deng et al. proposed Support Distance Error (SDE), computing the marginal gaps on the edges of a prediction and its targeted ground truth~\cite{deng2021revisiting}. Their proposal, however, is tailored for a point cloud-based object representation in the BEV plane. Similarly, Mori et al. considered several safety-oriented adaptations for existing evaluation protocols, e.g., by redefining the shortest distance between two bounding boxes~\cite{mori2024safety}. They also derived pass/fail criteria based on human perception performance. While the analysis with human benchmarks has been shown feasible, some assumptions may be overly complicated, as reported in the work itself. Henceforth, we address localization constraints that are simplistic and general for 3D object detectors using the typical bounding box-based representation.
	
	Lastly, there have also been interesting results that compensate for the imperfection of learning-based models. For example, De Gerancey et al. leveraged conformal prediction to post-process the predicted bounding boxes and obtain a statistical performance guarantee in ground truth alignment~\cite{degrancey2022object}. Similarly, Schuster et al. analytically derived a minimum multiplicative factor for enlarging the predictions so as to properly enclose the ground truths~\cite{schuster2022unaligned}. While these results focus on the 2D image plane, our work distinctively targets 3D object detection, which is more directly linked with AD applications and improves the models in a continual learning fashion.

	\section{Safety-Oriented Spatial Constraints}
	
	\begin{figure}
		\centering
		\includegraphics[width=0.98\linewidth]{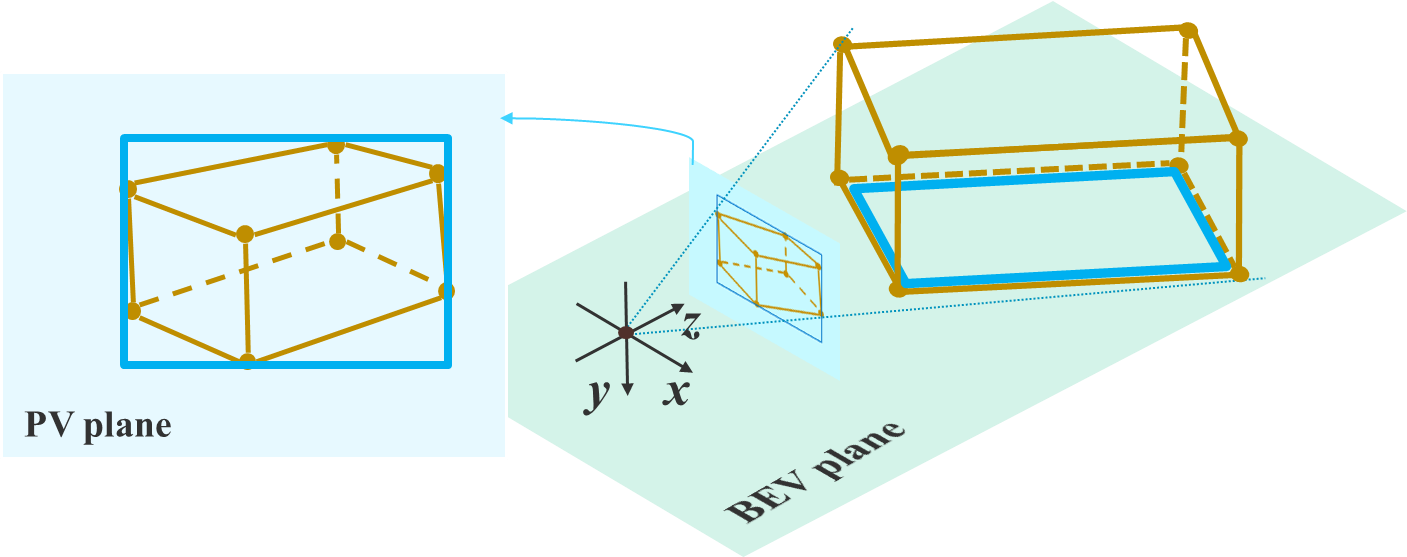}
		\caption{We transform 3D bounding boxes (BBs) onto the PV and BEV planes for establishing USC and corresponding quantitative measures. Gold: original BB; Blue: converted PV or BEV BB.}
		\label{fig:projections}
	\end{figure}
	
	To iterate, using the standard upright 3D bounding box representation~\cite{geiger2012are,caesar2020nuscenes}, our goal is to formulate uncompromising spatial constraints (USC) that ensure a given prediction~$\mathbf{P}$ fully covers the ground truth~$\mathbf{G}$ when seen by the AV. Technically, as Fig.~\ref{fig:projections} depicts, we transform them into the PV and BEV planes, which are easier to handle than the direct 3D world. In the following Sec.~\ref{subsec:pv}, we first address PV, which naturally reflects driving situations. Then, as PV foregoes the important depth information, we consider additional BEV constraints in Sec.~\ref{subsec:bev}. Finally, Sec.~\ref{subsec:consolidation} consolidates both PV and BEV constraints for comprehensive model evaluation.

	\subsection{Box Enclosure in the PV Plane}
	\label{subsec:pv}
	
	To formulate the spatial constraint in the PV plane, we first have to take a perspective projection for the prediction $\mathbf{P}$ and the ground truth $\mathbf{G}$. This can be done straightforwardly for camera-based object detectors with the camera parameters and images. As for lidar-based models, one may borrow the camera system in the sensor suite or simply apply a pinhole-based perspective projection using a focal length $f = 1$. This is because our following analysis only considers the relative position and size of the projected prediction and ground truth. As long as the same perspective projection (with any $f$) is applied to both the prediction $\mathbf{P}$ and the ground truth $\mathbf{G}$, we can examine their relation in the PV plane. To elaborate, for a point $p\defeq(x_p, y_p, z_p)$ in the 3D world, one can obtain its coordinates $(a, b)$ in the PV plane with the projection formula~\cite{hartley2004multiple}, written as:
	\begin{equation}
		\begin{bmatrix}
			a \\ 
			b \\
			f
		\end{bmatrix}
		=
		\frac{f}{z_p}
		\begin{bmatrix}
			x_p \\
			y_p \\
			z_p 
		\end{bmatrix}.
	\end{equation}
	The magnitude of the coordinates $(a, b)$ depend on the focal length $f$, but, as mentioned, it will be the same for all points in the prediction $\mathbf{P}$ and the ground truth $\mathbf{G}$.
	
	Practically, as shown in Fig.~\ref{fig:projections}, one only needs to consider the extreme points, i.e., the eight corners, of a 3D bounding box when performing the PV projection~\cite{liu2019deep}. As the eight corners result in eight points in the PV plane, an axis-aligned bounding box can finally be constructed by finding the minimum and maximum coordinates in the vertical and horizontal axes respectively. In this way, we obtain the projected prediction $\mathbf{P}^\mathcal{PV}$ and ground truth $\mathbf{G}^\mathcal{PV}$.

\begin{figure}[]
		\centering
		\includegraphics[width=0.8\linewidth]{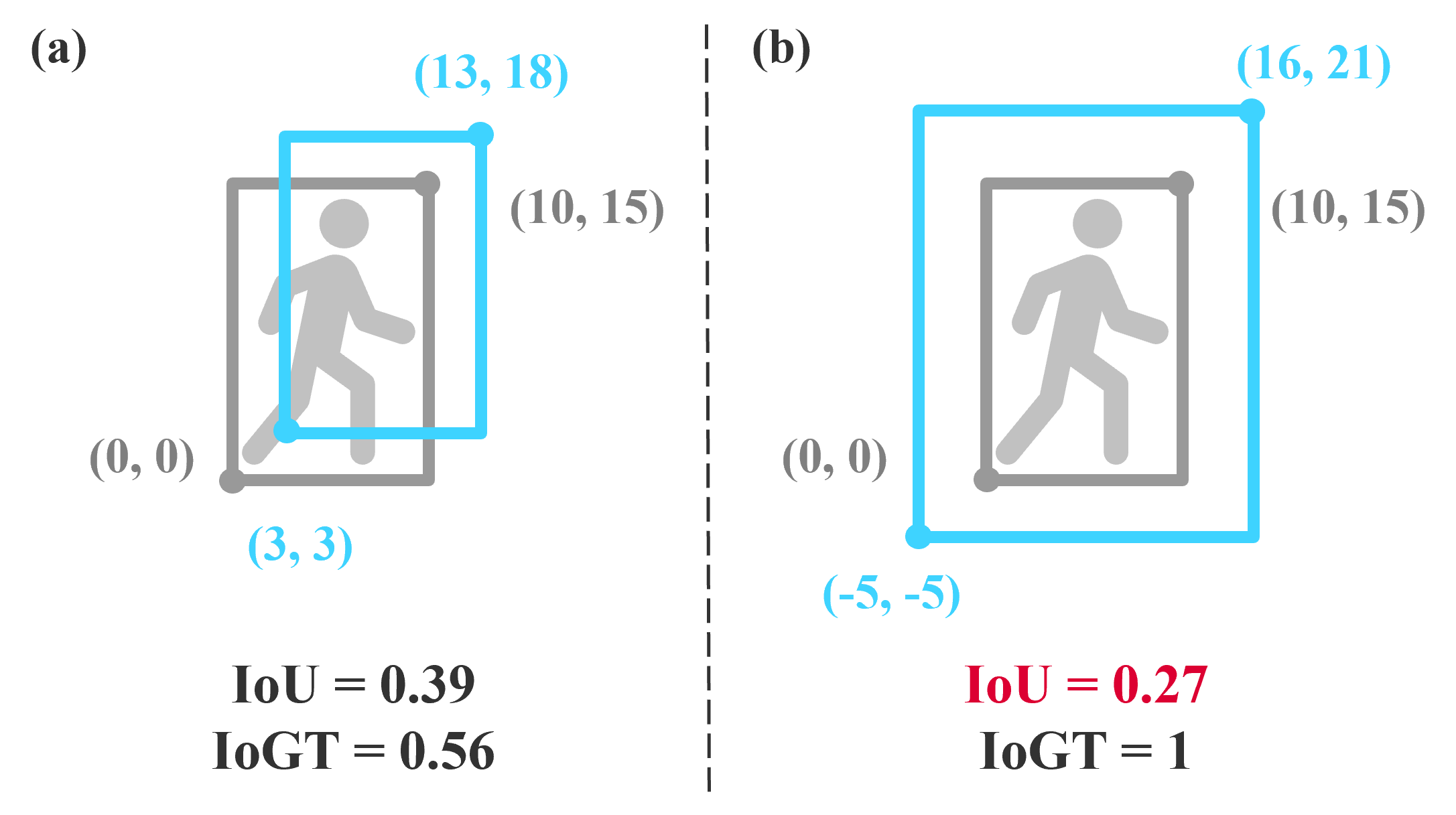}
		\caption{A comparison of IoU and IoGT using the same ground truth~\textit{(gray)} and different predictions~\textit{(blue)} in \textbf{(a)} and \textbf{(b)}. The prediction in \textbf{(b)} should be indicated as better than the one in \textbf{(a)} when considering collision avoidance. However, the IoU in \textbf{(b)} is lower, falling short to reflect the notion of ``enclosure".}
		\label{fig:IoGT}
	\end{figure}

	Similar to the prior work~\cite{degrancey2022object,schuster2022unaligned}, we now formulate a direct constraint in the PV plane that requires the prediction $\mathbf{P}^\mathcal{PV}$ to fully enclose the ground truth $\mathbf{G}^\mathcal{PV}$, i.e.,
	\begin{equation}
		\Pi^\mathcal{PV} \defeq \mathbf{G}^\mathcal{PV} \subset \mathbf{P}^\mathcal{PV}.
		\label{eq:pv}
	\end{equation}
	Intuitively, without a proper enclosure, the AV risks a higher chance of bumping into the object on its sides. For a quantitative characterization, however, the previous work~\cite{degrancey2022object,schuster2022unaligned} adopted the IoU function in 2D. On the contrary, we find Intersection-over-Ground-Truth (IoGT) more suitable here: 
	\begin{equation}
		\mathsf{IoGT}(\mathbf{P}^\mathcal{PV}, \mathbf{G}^\mathcal{PV}) \defeq \frac{\mathsf{Area}(\mathbf{P}^\mathcal{PV} \cap \mathbf{G}^\mathcal{PV})}{\mathsf{Area}(\mathbf{G}^\mathcal{PV})},
		\label{eq:iogt}
	\end{equation}
	with $\mathsf{Area}(\cdot)$ computing the area of a given polygon. Fig.~\ref{fig:IoGT} illustrates our rationale. Concretely, while IoU indicates the degree of alignment, IoGT reflects the degree of enclosure and better serves the purpose of the PV constraint. In addition, IoGT maximizes at $1$ when the prediction $\mathbf{P}^\mathcal{PV}$ fully encloses the ground truth $\mathbf{G}^\mathcal{PV}$.
	
	Now, with the PV constraint and its quantitative measure formulated, there is another risk that may be neglected: Even when $\Pi^\mathcal{PV}$ in Eq.~\eqref{eq:pv} is satisfied, it is still possible that in the 3D world, the ground truth $\mathbf{G}$ is not fully covered by the prediction $\mathbf{P}$ from the AV's location. For instance, the prediction $\mathbf{P}$ might be farther from the AV yet much larger than the ground truth $\mathbf{G}$, so that after the perspective projection, $\mathbf{P}^\mathcal{PV}$ indeed encloses $\mathbf{G}^\mathcal{PV}$. To prevent such cases, we need to consider the depth information or, more specifically, distance underestimation in the BEV plane.

	\subsection{Distance Underestimation in the BEV Plane}
	\label{subsec:bev}
	
	As motivated, a rule of thumb in the BEV plane is that the prediction should be no farther than the ground truth. We formulate such a rule and its quantitative measure in the following.
	
	First, as in the PV plane, we need the projected bounding boxes in the BEV plane. To such end, we perform an orthographic projection on the prediction~$\mathbf{P}$ and the ground truth~$\mathbf{G}$. Since we have assumed upright 3D bounding boxes, an orthographic projection simply boils down to taking the rectangle on the BEV plane, as Fig.~\ref{fig:projections} shows. In the following, we denote the projected prediction as~$\mathbf{P}^\mathcal{BEV}$ and ground truth as~$\mathbf{G}^\mathcal{BEV}$.
	
	For assessing the prediction $\mathbf{P}^\mathcal{BEV}$, the prior art~\cite{mori2024safety} has suggested considering the closest-point distance rather than the center distance~\cite{caesar2020nuscenes}, especially in safety-critical situations. We take the same rationale and denote the point closest to the AV in the ground truth as $v_\mathbf{G}^c$ and the one in the prediction as $v_\mathbf{P}^c$. Then, the requirement of closest-point distance underestimation can be specified as:
	\begin{equation}
		\Pi^\mathcal{BEV}_c \defeq || v_\mathbf{P}^c || \leq || v_\mathbf{G}^c||,
	\end{equation}
	where $||\cdot||$ here denotes the point distance to the origin.

\begin{figure}[]
		\centering
		\includegraphics[width=0.69\linewidth]{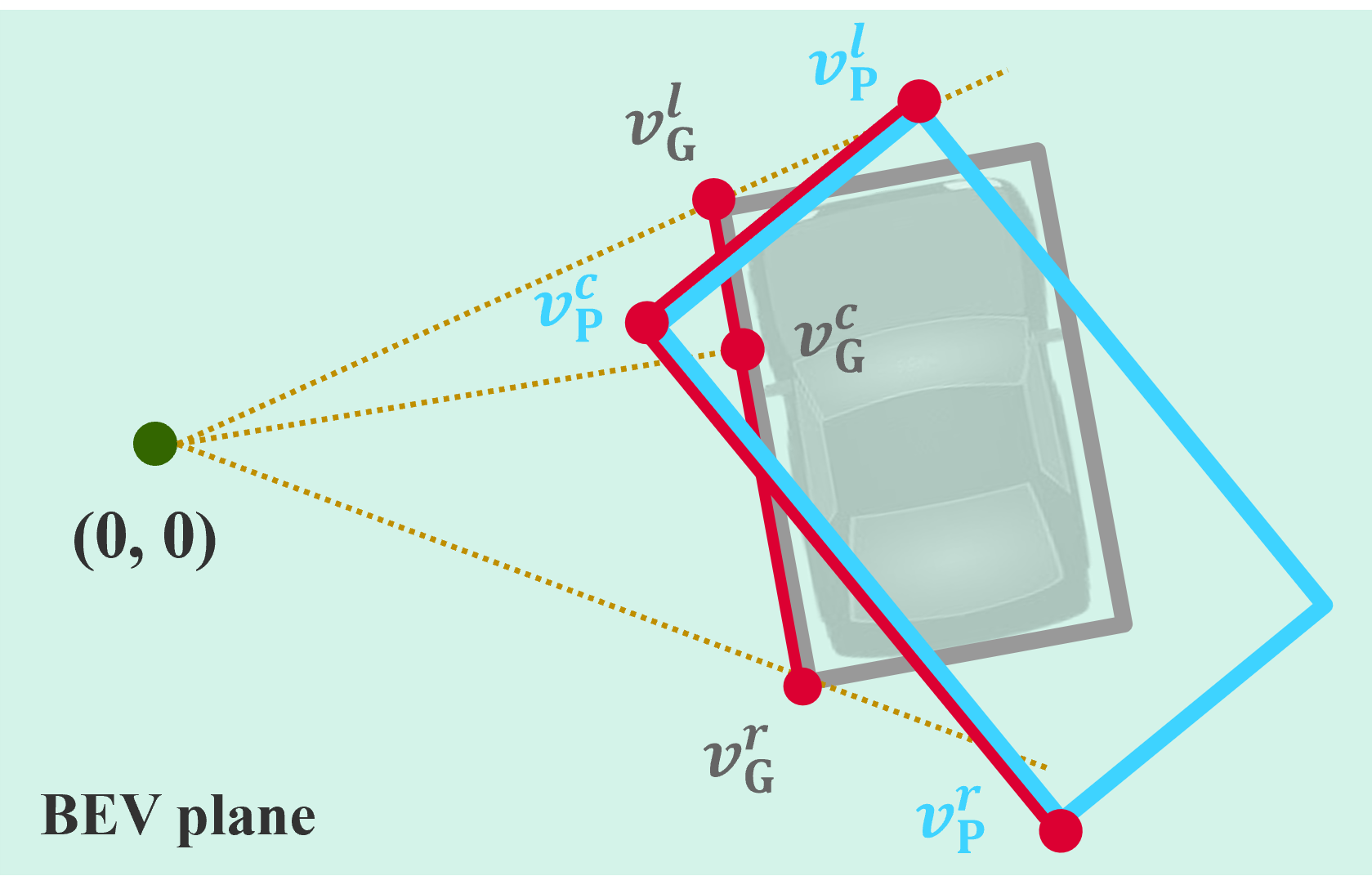}
		\caption{An example showing the risk where with the prediction's closest point $v_\mathbf{P}^c$ placed closer than the ground truth's $v_\mathbf{G}^c$, parts of the ground truth may still be exposed. Hence, we suggest taking into account the entire AV-facing sides, which are defined by the right-most and left-most corners from the AV's perspective. }
		\label{fig:bev}
	\end{figure}

	That said, as depicted in Fig.~\ref{fig:bev}, we note a pitfall where $\Pi^\mathcal{BEV}_c$ is satisfied but some part of the ground truth $\mathbf{G}^\mathcal{BEV}$ is still exposed. For full coverage over the ground truth $\mathbf{G}^\mathcal{BEV}$, one has to consider the entire side(s) facing the AV. Hence, we utilize the extreme points of $\mathbf{P}^\mathcal{BEV}$ and $\mathbf{G}^\mathcal{BEV}$ from the AV's perspective and further ensure there is no intersection in these sides. Formally, we now define the BEV constraint as:
	\begin{equation}
		\begin{split}
			\Pi^\mathcal{BEV} & \defeq  \Pi^\mathcal{BEV}_c \; \wedge \\ 
			& \neg \bot(\{ \overline{v_\mathbf{P}^c v_\mathbf{P}^r}, \overline{v_\mathbf{P}^c v_\mathbf{P}^l}, \overline{v_\mathbf{G}^c v_\mathbf{G}^r}, \overline{v_\mathbf{G}^c v_\mathbf{G}^l} \}),
		\end{split}
		\label{eq:bev}
	\end{equation}
	where $v_\mathbf{G}^r$ and $v_\mathbf{G}^l$ denote the right-most and left-most corners in the ground truth $\mathbf{G}^\mathcal{BEV}$ from the AV's perspective (similarly for the prediction $\mathbf{P}^\mathcal{BEV}$), and $\bot(\cdot)$ is a function that checks if a given set of line segments has an intersection, excluding overlapping endpoints. In practice, the function $\bot(\cdot)$ can be implemented by the Sweep Line Algorithm~\cite{souvaine2005line}.

	For the quantitative measure in the BEV plane, we observe that, in general, the farther the prediction is placed from the AV in comparison to the ground truth, the larger a portion of the ground truth is exposed. Therefore, we leverage the defined representative points (i.e., the closest, left-most, and right-most points from the AV's perspective) and formulate an average distance ratio (ADR) as follows:
	\begin{equation}
		\begin{split}
			\mathsf{ADR}(\mathbf{P}^\mathcal{BEV}, & \mathbf{G}^\mathcal{BEV})  \defeq \\ 
			& \left( \prod_{ i = \{c, r, l \} } \frac{ || v_\mathbf{G}^i|| }{ \max ( || v_\mathbf{P}^i||, || v_\mathbf{G}^i|| )} \right) ^ { (1/3)}.
		\end{split}
		\label{eq:adr}
	\end{equation}
	Essentially, we compute individually how far the prediction's representative points are from the AV compared to their ground truth counterparts and aggregate the point-wise ratio by taking the geometric mean. Here, using the geometric mean instead of other central tendencies, such as the arithmetic mean, gives more robust results, e.g., in the case of a large prediction~\cite{fleming1986how}. Lastly, similar to IoGT, which maximizes when the prediction arrives at a ``safe" state, ADR also saturates at $1$ if the prediction's representative points are all closer than their corresponding ground truth points.

	\subsection{Consolidation for Model Evaluation}
	\label{subsec:consolidation}
	
	So far, we have discussed the PV and BEV constraints as well as their quantification separately. To generate a verdict for a prediction $\mathbf{P}$ with the ground truth $\mathbf{G}$ in the 3D world, we consolidate the uncompromising spatial constraints (USC) from Eq.~\eqref{eq:pv} and Eq.~\eqref{eq:bev} as:
	\begin{equation}
		\Pi_\mathsf{USC} \defeq \Pi^\mathcal{PV} \wedge \Pi^\mathcal{BEV}.
	\end{equation}
	Thereby, we obtain a qualitative indicator checking if the prediction $\mathbf{P}$ fully covers the ground truth $\mathbf{G}$ when seen from the AV. In addition, with the formulated PV and BEV quantitative measures in Eq.~\eqref{eq:iogt} and Eq.~\eqref{eq:adr}, we can produce a scoring function as:
	\begin{equation}
		\mathsf{USC}(\mathbf{P}, \mathbf{G}) \defeq \mathsf{IoGT} \times \mathsf{ADR},
	\end{equation}
	where we omit the notation of the projected PV and BEV bounding boxes for simplicity. Generally, a prediction $\mathbf{P}$ gets a higher USC score if it covers the ground truth $\mathbf{G}$ better. Since both IoGT and ADR range from $0$ to $1$, the overall score also ranges over $[0,1]$.

	As seen, USC is designed as a TP measure to examine a pair of matched prediction $\mathbf{P}$ and ground truth $\mathbf{G}$. After examining the matched pairs for an object class, one can calculate an~\emph{average USC (AUSC)}. Then, summarizing the results from all object classes gives the~\emph{mean AUSC (mAUSC)}. Furthermore, to also reflect false-positive (FP) and false-negative (FN) predictions, we follow nuScenes' protocol~\cite{caesar2020nuscenes}, which incorporates already the recall-precision-based mAP metric, and simply extend mAUSC as follows:  
	\begin{equation}
		\mathsf{USC\text{-}NDS} = \frac{1}{2} [ \mathsf{NDS} + \mathsf{mAUSC} ].
	\end{equation}
	As such, \emph{USC-NDS} not only reflects the generic object detection performance but also characterizes our safety-oriented spatial constraints from the driving perspective. To see their efficacy, we shall use the consolidated AUSC, mAUSC, and USC-NDS for model evaluation in our experiments later.

	\section{Safety-Oriented Fine-Tuning}
	\label{sec:loss}
	
	Apart from examining model performance with the safety-oriented spatial constraints, we consider how to improve them via a loss function for continual learning. In particular, along with our analysis, we notice that an ultimate safety-oriented goal in AD is to encapsulate the ground truths with predictions. That is, while the spatial constraints tolerate certain errors and require a prediction to cover the ground truth when seen from the AV, we directly aim for proper containment with the 3D bounding boxes during optimization.
	
	It turns out that a proper containment can be modeled by the 3D version of the IoGT function, comparing volumes of the intersection~$\mathbf{P} \cap \mathbf{G}$ and ground truth~$\mathbf{G}$, similar to Eq.~\eqref{eq:iogt}. The better a prediction contains the ground truth, the higher its IoGT score. With full containment, IoGT maximizes at~$1$. Therefore, we compose an IoGT loss as:
	\begin{equation}
		L_{\mathsf{IoGT}} \defeq 1 - \mathsf{IoGT}(\mathbf{P}, \mathbf{G}).
	\end{equation}
	Nevertheless, the IoGT loss by itself does not provide a complete solution since it vanishes when the prediction over-approximates the ground truth. We consequently combine it with accuracy-guided loss functions such as the SmoothL1 loss ($L_{\mathsf{SmoothL1}}$)~\cite{liu2016ssd} and obtain a safety-oriented loss function:
	\begin{equation}
		L_{\mathsf{safety}} \defeq \lambda \cdot L_{\mathsf{SmoothL1}} + (1-\lambda) \cdot L_{\mathsf{IoGT}},
		\label{eq:safety_loss}
	\end{equation}
	where $\lambda \in \mathbb{R}: 0 < \lambda < 1$ is a balancing parameter. In practice, similar to the NN learning rate, $\lambda$ can be tuned during a learning session to favor accuracy in early episodes and improve safety-related performance at later stages.

	\section{Experiments}
	
	With the proposed USC-based metrics for model evaluation in Sec.~\ref{subsec:consolidation} and the safety-oriented loss function for model fine-tuning in Sec.~\ref{sec:loss}, we now conduct experiments to answer the following three research questions (RQs).
	\begin{itemize}
		\item RQ1: Given their generally good performances, how do state-of-the-art models perform in terms of the safety-oriented USC-based metrics?
		\item RQ2: Can the proposed loss function improve the model's performances, especially regarding the safety-oriented scores?
		\item RQ3: Ultimately, are USC-based metrics better performance indicators for system safety?
	\end{itemize}
	The RQs will be investigated in the following three sections respectively.

	\subsection{Model Evaluation}
	
	We now start with RQ1 and evaluate various models in terms of the existing accuracy-based mAP and NDS, as well as our USC-based metrics.

	\begin{figure}[]
		\centering
		\includegraphics[width=0.8\linewidth]{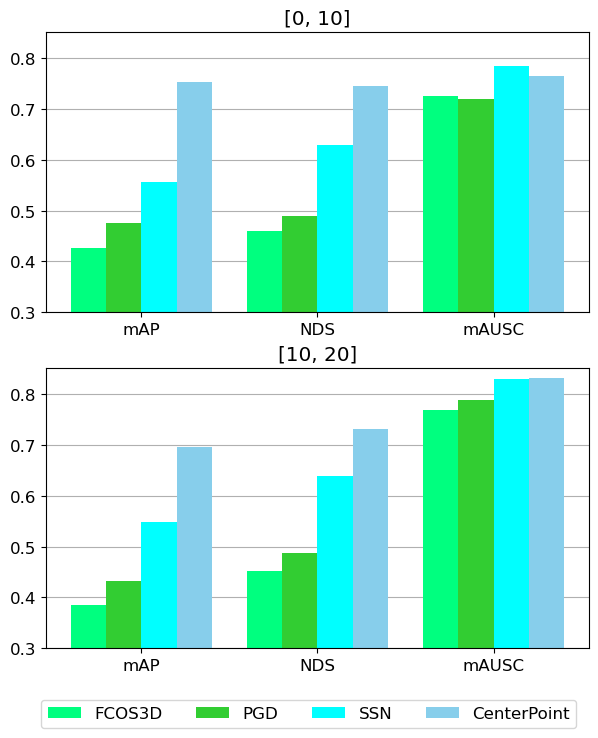}
		\caption{Range-based model evaluation results. The upper plot shows the scores of the four examined models for objects ranging in $[0, 10]$ meters, and the lower $[10, 20]$ meters. The camera-based FCOS3D~\cite{wang2021fcos3d} and PGD~\cite{wang2021probabilistic} are plotted in green, and the lidar-based SSN~\cite{zhu2020ssn} and CenterPoint~\cite{yin2021center} are in blue.}
		\label{fig:range_analysis}
	\end{figure}

	\subsubsection{Implementation}
	As introduced, we perform model evaluation using the nuScenes dataset~\cite{caesar2020nuscenes}. Given our safety-oriented considerations for collision avoidance, we make several adaptations to its evaluation protocol. First, we concentrate on objects within $20$ meters instead of $50$. Then, we separate the objects by their ranges to give a more detailed analysis. For objects within $0$ to $10$ meters, we tighten the TP threshold for center distances from $2$ to $1$ meters. Lastly, some object classes do not exist in specific ranges. For example, no construction vehicles are situated within $[0, 10]$ meters in all scenes. For such cases, the original nuScenes protocol assumes the worst scores, e.g., $0$ AP and full TP errors, leading to overly underestimated performance. Finding that sub-optimal, we do not account for missing classes in our experiments.
	
	For the object detection models, we employ the comprehensive MMDetection3D library~\cite{mmdet3d2020} We adopt the top-performing ones in the nuScenes benchmark~\cite{caesar2020nuscenes}, including camera-based FCOS3D~\cite{wang2021fcos3d} and PGD~\cite{wang2021probabilistic} as well as lidar-based SSN~\cite{zhu2020ssn} and CenterPoint~\cite{yin2021center}.

	\subsubsection{Results and Discussions}
	
	We plot the evaluation results of the four models in Fig.~\ref{fig:range_analysis}, where we observe the safety-oriented performance of the models does not necessarily follow the same trend as their accuracy-based performance. We elaborate on the observation with the following three points.
	
	First, while there are large accuracy gaps between camera-based and lidar-based models, the gaps in terms of mAUSC are much smaller. This likely shows that when it comes to localizing nearby objects and covering them with ``safe" enough predictions, the camera-based models are not as bad as indicated by the accuracy-based metrics. 
	
	Second, looking at the plot for $[0, 10]$ meters, the mAUSC score of PGD is actually lower than that of FCOS3D, despite it having higher mAP and NDS. A similar situation appears for the lidar-based CenterPoint and SSN. This finding resonates with the related work~\cite{deng2021revisiting}, pointing out that more accurate models may still exhibit safety concerns in short ranges.
	
	Third and lastly, from the $[0, 10]$ range to the $[10, 20]$ range, the mAP and NDS of the models tend to drop, while their mAUSC interestingly increase. This may be attributed to the error-tolerating principles when designing mAUSC. Essentially, the objects at a distance tend to be harder to localize, but it may be fine so long as the predictions cover the ground truths (as exemplified by Fig.~\ref{fig:example}). 
	
	Altogether, the model evaluation results using mAP, NDS, and our USC-based metrics demonstrate that state-of-the-art models may have different safety-oriented performance from what is generally perceived.

	\subsection{Model Fine-Tuning}

	\begin{table*}[]
		\begin{minipage}{\textwidth}
			\centering
			\caption{Quantitative results of fine-tuning the camera-based PGD model~\cite{wang2021probabilistic}. NDS and ATE are accuracy-oriented metrics from nuScenes' protocol~\cite{caesar2020nuscenes}, whereas mAUSC and AUSC are our safety-oriented metrics. Using the proposed $L_{\mathsf{safety}}$ not only achieves better safety-oriented scores but also maintains the accuracy.}
			\label{tab:quantitative_results}
			\begin{tabular}{l|ccc|cc|cc|cc}
				\hline
				& \multicolumn{1}{c|}{}                           & \multicolumn{1}{c|}{}                      & \multicolumn{1}{c|}{}                                                  & \multicolumn{2}{c|}{Pedestrian}                                                   & \multicolumn{2}{c|}{Car}                             & \multicolumn{2}{c}{Truck}                                                   \\ \cline{5-10} 
				\multirow{-2}{*}{Model}   & \multicolumn{1}{c|}{\multirow{-2}{*}{Modality}} & \multicolumn{1}{c|}{\multirow{-2}{*}{NDS (\%)}} & \multicolumn{1}{c|}{\multirow{-2}{*}{mAUSC (\%)}} & \multicolumn{1}{c|}{ATE ($\downarrow$)}            & AUSC ($\uparrow$)                                 & \multicolumn{1}{c|}{ATE ($\downarrow$)}            & AUSC ($\uparrow$)           & \multicolumn{1}{c|}{ATE} ($\downarrow$)                                   & AUSC ($\uparrow$)          \\ \hline
				PGD~\cite{wang2021probabilistic}                       & \multicolumn{1}{c|}{Camera}                     & \multicolumn{1}{c|}{50.76}                 & \multicolumn{1}{c|}{75.35}                  & \multicolumn{1}{c|}{0.439}          & 0.728                                 & \multicolumn{1}{c|}{0.326}          & 0.811          & \multicolumn{1}{c|}{0.498}                                 & 0.803          \\ \hdashline
				+$L_\mathsf{SmoothL1}$~\cite{liu2016ssd} & \multicolumn{1}{c|}{Camera}                     & \multicolumn{1}{c|}{50.68 (-0.2\%)}                 & \multicolumn{1}{c|}{75.69 (+0.5\%)}                   & \multicolumn{1}{c|}{0.433}          & 0.729                                 & \multicolumn{1}{c|}{0.325}          & 0.804          & \multicolumn{1}{c|}{0.497}                                 & 0.794          \\
				+$L_\mathsf{safety}$ (ours)       & \multicolumn{1}{c|}{Camera}                     & \multicolumn{1}{c|}{\textbf{50.88} (+0.2\%)}        & \multicolumn{1}{c|}{\textbf{78.54} (+4.2\%)}          & \multicolumn{1}{c|}{0.434}          & 0.732                                 & \multicolumn{1}{c|}{0.324}          & 0.813          & \multicolumn{1}{c|}{0.503}                                 & 0.805          \\ \hline
			\end{tabular}
		\end{minipage}
	\end{table*}

	\begin{figure*}
		\centering
		\begin{minipage}[]{0.48\textwidth}
			\begin{minipage}[b]{0.48\linewidth}
				\includegraphics[width=\linewidth]{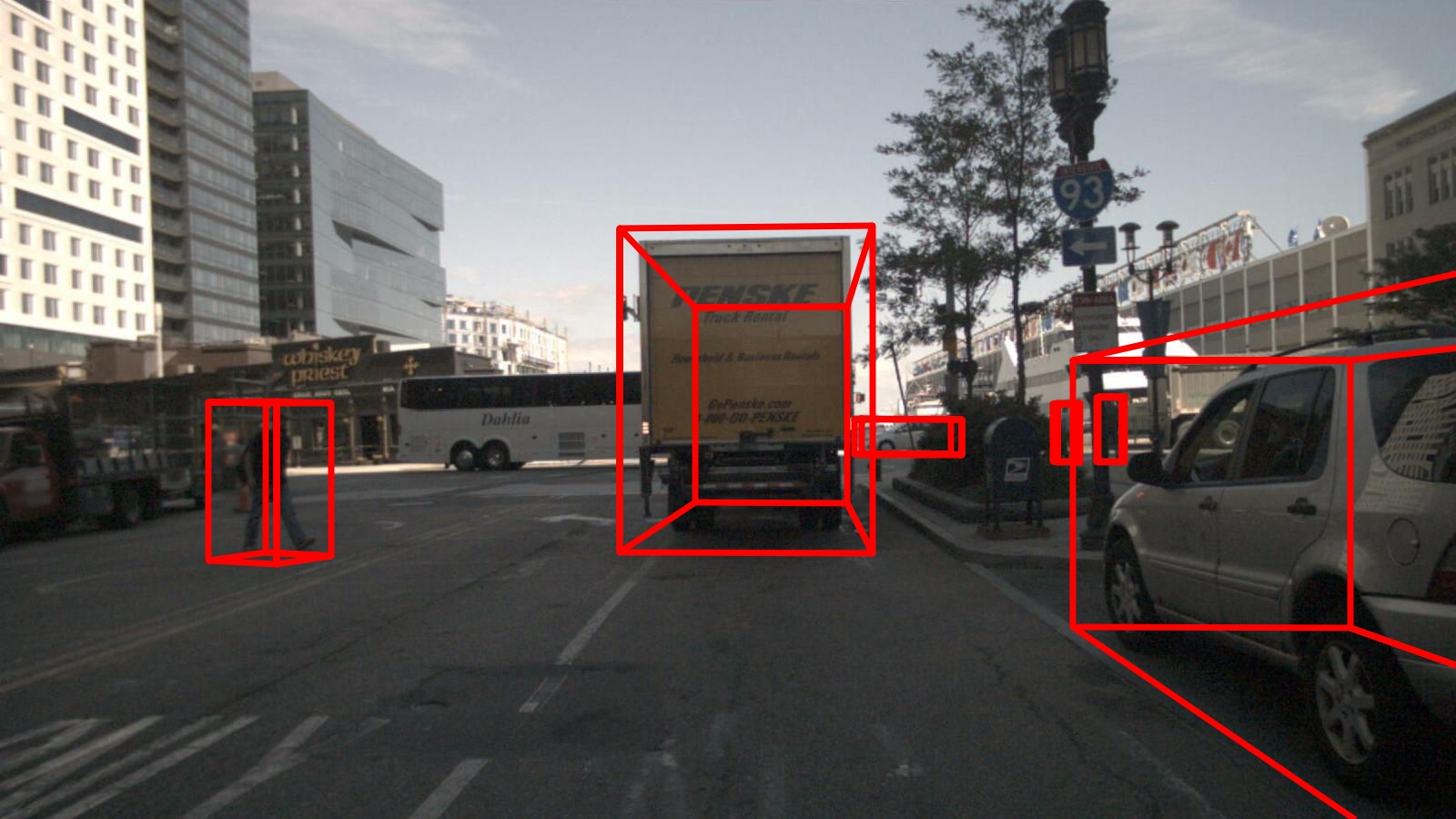}
				\vspace{1mm}
				\includegraphics[width=\linewidth]{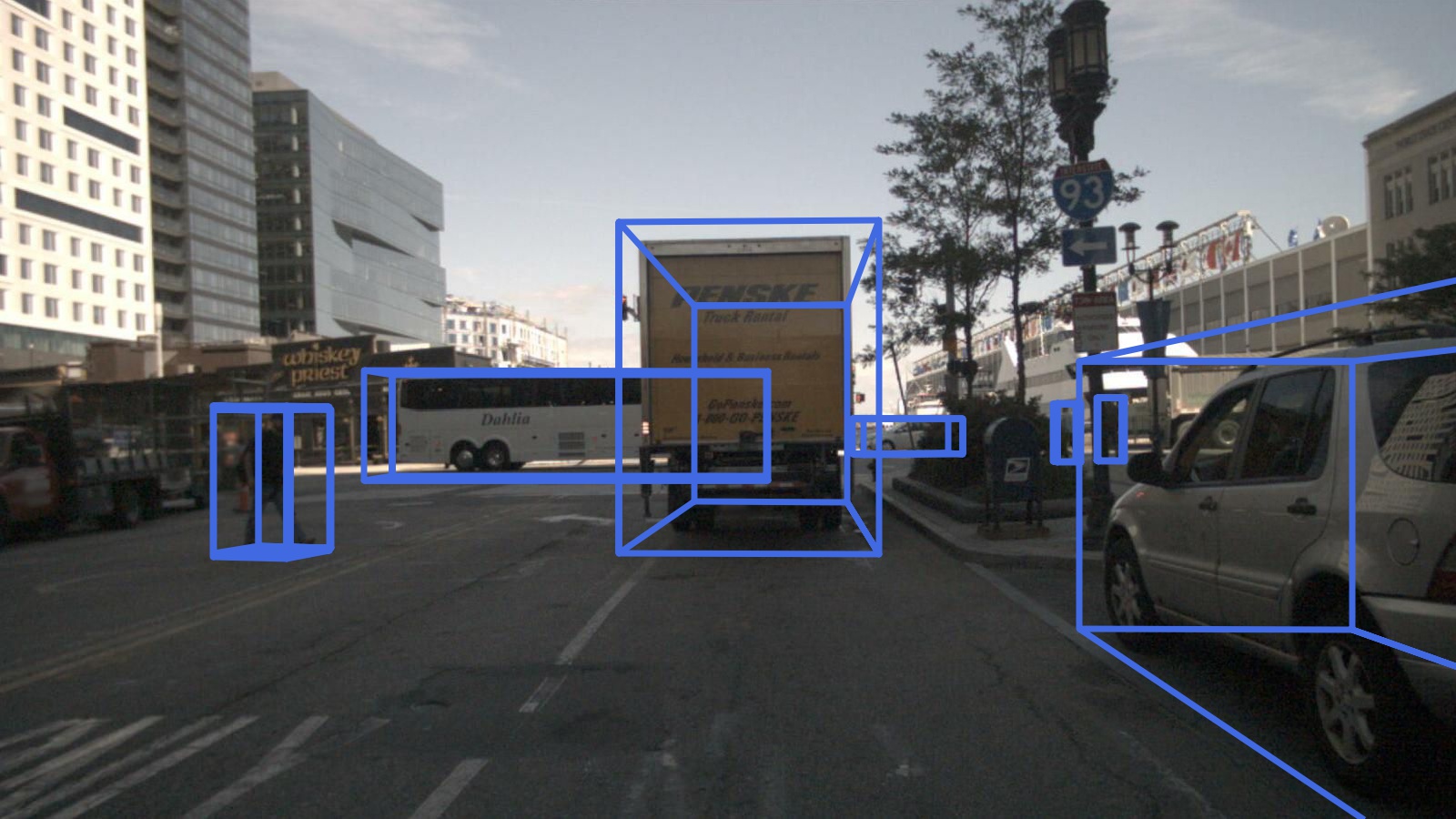}    
			\end{minipage}
			\includegraphics[width=0.48\linewidth]{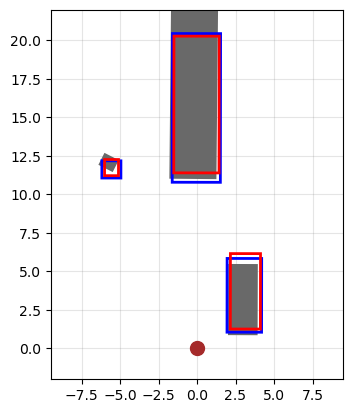}
		\end{minipage}
		\hspace{1mm}
		\begin{minipage}[]{0.48\textwidth}
			\begin{minipage}[b]{0.48\linewidth}
				\includegraphics[width=\linewidth]{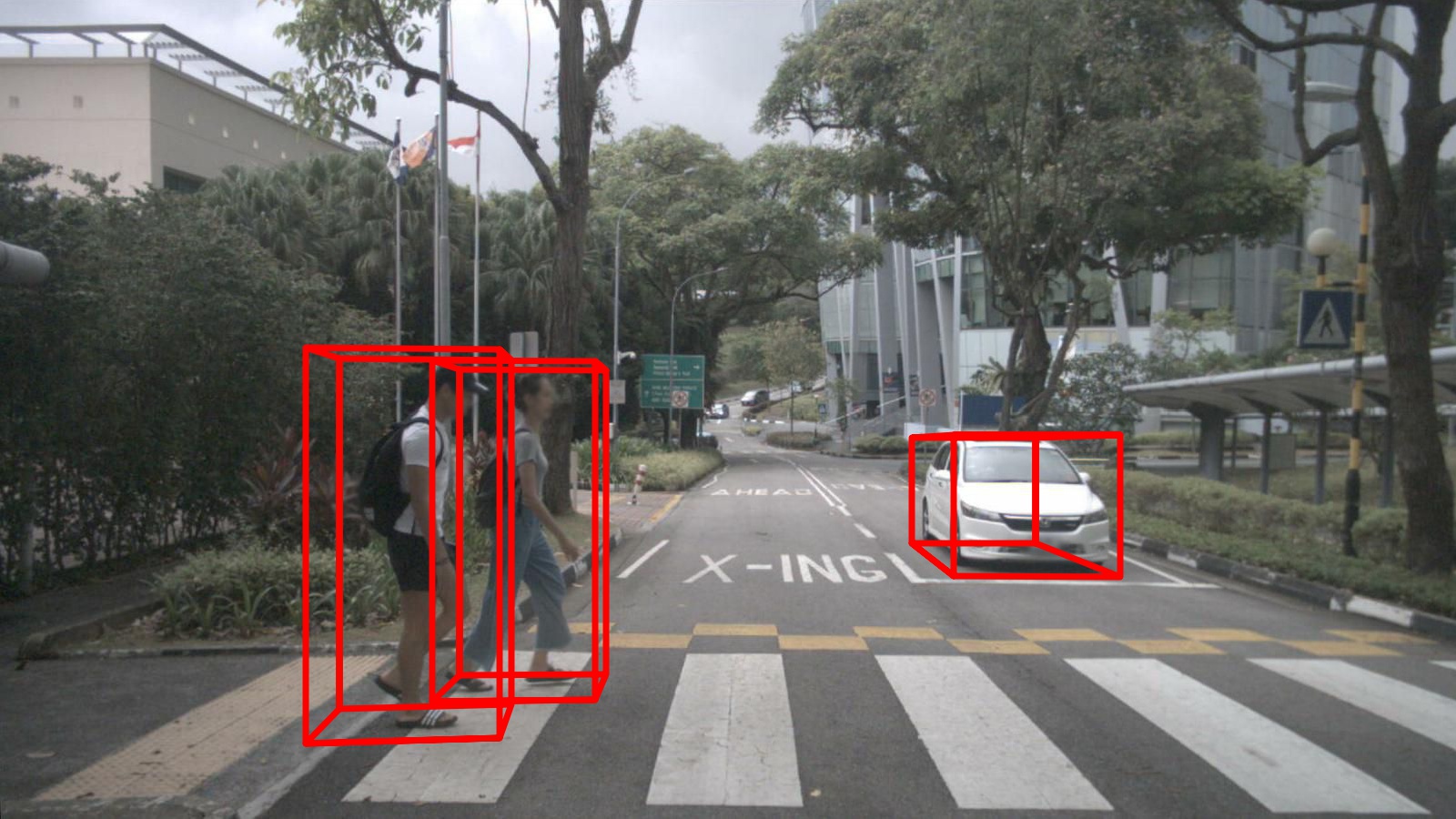} 
				\vspace{1mm}
				\includegraphics[width=\linewidth]{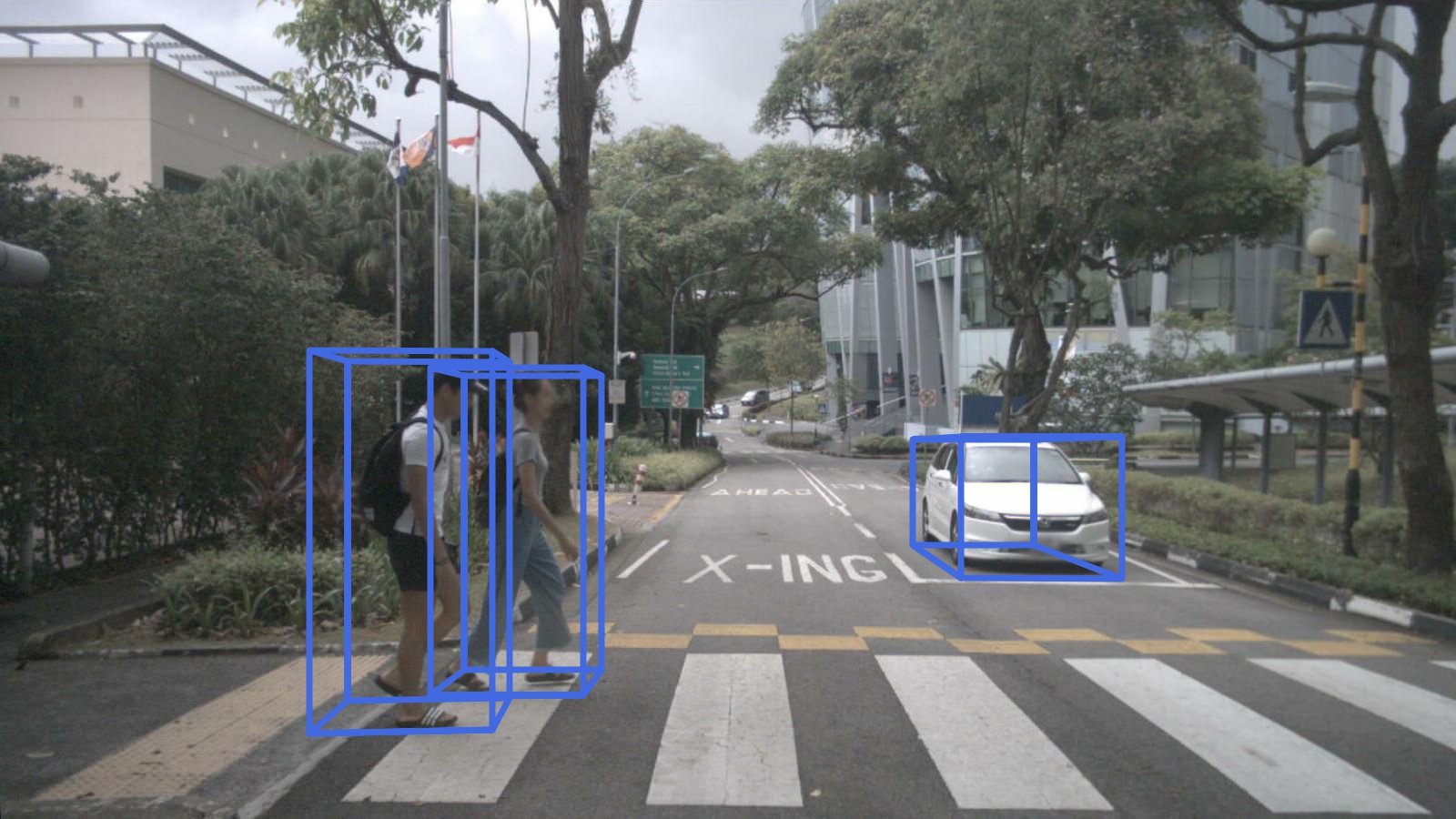}    
			\end{minipage}
			\includegraphics[width=0.48\linewidth]{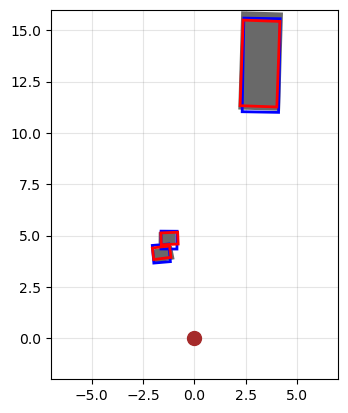}
		\end{minipage}
		\caption{Quantitative results of fine-tuning camera-based PGD model~\cite{wang2021probabilistic}. For each of the two scenes, the upper left image shows red predictions based on~$L_{\mathsf{SmoothL1}}$~\cite{liu2016ssd}, the lower left image shows blue predictions based on our~$L_{\mathsf{safety}}$, and on the right is a BEV projection of all predictions alongside the ground truths in gray. In general, the blue predictions contain the ground truths to a larger extent while staying aligned with the ground truths.}
		\label{fig:qualitative_results}
	\end{figure*}

	In this section, we continue with RQ2. In particular, considering the potential safety concern of PGD discovered in RQ1, we apply the safety-oriented loss function for it.
	
	\subsubsection{Implementation}
	
	Since the PGD object detector originally outputs a unique image-based representation during the learning phase~\cite{wang2021probabilistic}, we append to it a transformation function and obtain the 3D bounding box representation. Then, together with the annotated ground truths, the IoGT loss and safety-oriented loss (with $\lambda$ set to $0.8$) can be calculated. We run five fine-tuning sessions for the best PGD checkpoint provided on MMDetection3D~\cite{mmdet3d2020} using the same learning configurations as the original PGD work, e.g., NN learning rate and data augmentation policy. For comparison, we also run five sessions with the accuracy-guided loss function $L_{\mathsf{SmoothL1}}$. All sessions are run with a batch size of six samples for six epochs of a subset shared by the nuScenes benchmark~\cite{caesar2020nuscenes}. Each session takes roughly two hours on an Nvidia RTX A6000.

	\subsubsection{Results and Discussions}

	We present in Tab.~\ref{tab:quantitative_results} the fine-tuning results, which are the average of the five sessions for the $L_{\mathsf{SmoothL1}}$-based models and the $L_{\mathsf{safety}}$-based models respectively. We also provide qualitative samples from two scenes in Fig~\ref{fig:qualitative_results}.
	
	Interestingly, as Tab.~\ref{tab:quantitative_results} shows, the proposed loss function not only improves the model's safety-oriented performance in terms of mAUSC but also maintains, or even slightly increases, the accuracy-based NDS. This likely stems from the explicit usage of 3D IoGT in the loss function defined by Eq.~\eqref{eq:safety_loss}, which aims not only at full coverage from the AV's perspective but at proper containment of the ground truths, thereby keeping the predictions aligned at the same time. The effect is visualized in the qualitative samples, where most of the blue predictions contain the ground truths to a larger extent and are aligned well.
	
	A similar trend can be observed regarding the class-specific results: The safety-oriented models achieve better AUSC scores while maintaining similar average translation error (ATE). For example, corresponding to the higher AUSC, the pedestrians are better contained by the blue predictions in the right scene of Fig.~\ref{fig:qualitative_results}. 
	
	In summary, the proposed loss function $L_{\mathsf{safety}}$ does improve model performances, especially regarding the safety-oriented scores. Although the improvement might be marginal, such differences may play an important role in safety-critical situations. For further enhancement, nonetheless, one may consider adopting complementary approaches such as box enlarging in a post-processing fashion~\cite{degrancey2022object,schuster2022unaligned}. Moreover, we leave a full examination and reconciliation of the potentially competing accuracy- and safety-oriented objectives in $L_{\mathsf{safety}}$ as future work.

	\subsection{Closed-Loop Validation}
	
	Finally, in this section, we explore RQ3 and validate the proposed safety-oriented USC-based metrics with a system-level simulation. In principle, to be indicative and effective, a model-level metric should have a good correlation with the actual system-level testing result.

	\subsubsection{Implementation}
	
	We run all the baseline and fine-tuned camera-based object detectors on an AV equipped with an automatic braking function (AEB) in the Prescan simulator. For each of the twelve models (two baselines and ten fine-tuned), we repeat 100 simulations. Then, we compute the Pearson correlation coefficients between the model-level metrics and the AV collision rates.
	
\begin{figure}[]
		\centering
		\includegraphics[width=0.7\linewidth]{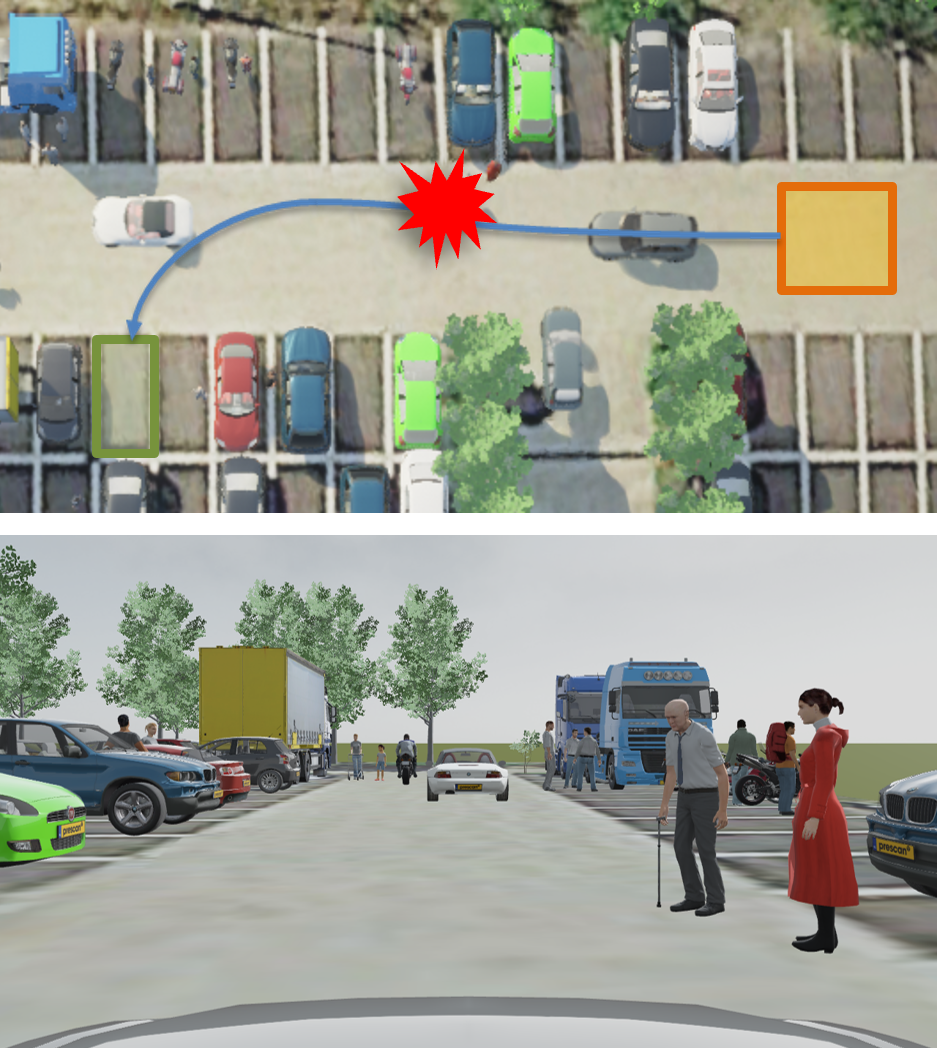}
		\caption{The system-level simulation setup: \textit{(Top)} The gray AV equipped with the object detector and AEB function is commissioned to drive from the yellow hand-off zone to the green parking slot. \textit{(Bottom)} During the mission, the pedestrian at the back (e.g., an old man here) crosses the driveway and causes a collision potentially.}
		\label{fig:sim}
	\end{figure}

	To briefly explain, in every simulation, the AV's mission is to perform valet parking by driving along a designated path from a hand-off zone to a parking slot. During the mission, a pedestrian will cross the driveway, thereby testing the object detection and AEB functions of the AV. The AEB function operates based on a simple rule: If an object is detected within the braking range of the AV, it will take a full brake. Hence, whether a collision occurs depends mainly on the object detector's performance. Lastly, for more diverse testing, we alter the pedestrian type, speed, crossing timing, the AV's speed, and environmental conditions such as illumination. Fig.~\ref{fig:sim} illustrates the setup. Readers can check our demonstration video~\cite{liao2023demo} or Sec. 4.2 of an overviewing project paper~\cite{bensalem2023continuous} for more details regarding the AV's simulation and testing.
	
	\subsubsection{Results and Discussions}
	
	\begin{table}[b]
		\caption{Absolute Pearson correlation coefficients between model-level metrics and the simulated system-level collision rates.}
		\label{tab:correlation}
		\centering
		\begin{tabular}{lc}
			\textbf{Metric}                       & \textbf{Correlation} \\ \hline
			mAP (mean Average Precision)          & 0.699                \\
			NDS (NuScenes Detection Score)        & 0.806                \\
			mAUSC (mean Average USC)              & 0.814                \\
			USC-NDS (an average of the above two) & \textbf{0.925}       \\ \hline
		\end{tabular}
	\end{table}
	
	Generally, we see the fine-tuned PGD models with higher safety-oriented scores lead to fewer simulation collisions. Tab.~\ref{tab:correlation} summarizes the results with the Pearson correlation coefficients. For clarity, we report the absolute values so that the higher the value, the more indicative the model-level metric. 
	
	Compared to the conventional mAP, NDS already reflects the system-level performance well. This is similarly concluded in the recent validation work~\cite{schreier2023offline}. Nonetheless, the proposed mAUSC outperforms NDS, and by combining the two, one can attain the most indicative model-level metric for the system-level collision rates. 
	
	To reflect on the results, we find that in most simulations, the pedestrians are detected (as in Fig.~\ref{fig:qualitative_results}), and mAUSC successfully characterizes the safety-oriented properties in these TP cases. Still, there are cases where the pedestrians are not detected at all or the predictions are too distant from the pedestrians. Considering these cases, the well-rounded USC-NDS metric, which also reflects such FNs and FPs, is indeed the most suitable proxy for system-level performance.

	\section{Conclusion}
	
	In this work, observing the urge for appropriate performance indicators for 3D object detectors, we formulated uncompromising spatial constraints (USC) and corresponding quantitative metrics based on a safety point of view. Concretely, given a prediction $\mathbf{P}$ and its targeted ground truth $\mathbf{G}$, the proposed USC requires the prediction $\mathbf{P}$ to cover the ground truth $\mathbf{G}$ when seen from the AV. Such a requirement foregoes the likely unachievable perfect detection demanded by conventional accuracy-based metrics such as mAP. We conducted extensive experiments with the nuScenes dataset and a closed-loop AEB system simulation. The results demonstrated the efficacy of USC-based metrics in finding safety-oriented and more dependable models as well as its potential to gear existing models towards better performances beyond pure accuracy. 
	
	On a broader scope, our investigation joins the recent efforts in AD safety engineering and puts forth a promising proposal for 3D object detectors. To proceed, there are several open directions. First, with the direct safety-oriented criterion, we plan to employ analytical or statistical approaches to get a reasonable performance threshold for deployment. Second, one may extend USC's usage in facilitating more advanced motion planning algorithms for smoother and safer maneuvers. Last but not least, how to adapt USC for run-time monitoring constitutes another interesting avenue.

	\balance
	\bibliographystyle{IEEEtran}
	\bibliography{ref}
	
\end{document}